\documentclass[sigconf]{acmart}
\AtBeginDocument{%
  }
\renewcommand\footnotetextcopyrightpermission[1]{} % removes footnote with conference information in first column

\setcopyright{none}
\usepackage{array}
\usepackage{booktabs}
\usepackage{multirow}
\usepackage{makecell}
\begin{document}

\title{ARIW-Framework: Adaptive Robust Iterative Watermarking Framework}

\author{Shaowu Wu}
\email{wushw25@mail2.sysu.edu.cn}
\affiliation{%
  \institution{School of Computer Science and
  	Engineering,\\Sun Yat-sen University}
  \city{Guangzhou}
  \state{}
  \country{China}
}

\author{Liting Zeng}
\email{zenglt9@mail2.sysu.edu.cn}
\affiliation{%
  \institution{School of Computer Science and
  	Engineering,\\Sun Yat-sen University}
  \city{Guangzhou}
  \country{China}}

\author{Wei Lu}
\authornote{Corresponding authors.}
\email{luwei3@mail.sysu.edu.cn}
\affiliation{%
  \institution{School of Computer Science and
  	Engineering,\\Sun Yat-sen University}
  \city{Guangzhou}
  \country{China}
}

\author{Xiangyang Luo}
\email{luoxy\_ieu@sina.com}
\affiliation{%
 \institution{State Key Laboratory of Mathematical
 	Engineering and Advanced Computing}
 \city{Zhengzhou}
 \state{}
 \country{China}}

%\author{Jiantao Zhou}
%\email{caoxiaochun@mail.sysu.edu.cn}
%\affiliation{%
%	\institution{Department of Computer and Information Science, University of Macau}
%	\city{Macau}
%	\state{}
%	\country{China}}
%\email{jtzhou@um.edu.mo}

%\author{Xiaochun Cao}
%\email{caoxiaochun@mail.sysu.edu.cn}
%\affiliation{%
%  \institution{School of Cyber Science and
 % 	Technology, Shenzhen Campus of Sun Yat-sen University}
 % \city{Shenzhen}
 % \state{}
 % \country{China}}

\renewcommand{\shortauthors}{Trovato et al.}

\begin{abstract}
  With the rapid rise of large models, copyright protection for generated image content has become a critical security challenge. Although deep learning watermarking techniques offer an effective solution for digital image copyright protection, they still face limitations in terms of visual quality, robustness and generalization. To address these issues, this paper proposes an adaptive robust iterative watermarking framework (ARIW-Framework) that achieves high-quality watermarked images while maintaining exceptional robustness and generalization performance. Specifically, we introduce an iterative approach to optimize the encoder for generating robust residuals. The encoder incorporates noise layers and a decoder to compute robustness weights for residuals under various noise attacks. By employing a parallel optimization strategy, the framework enhances robustness against multiple types of noise attacks. Furthermore, we leverage image gradients to determine the embedding strength at each pixel location, significantly improving the visual quality of the watermarked images. Extensive experiments demonstrate that the proposed method achieves superior visual quality while exhibiting remarkable robustness and generalization against noise attacks.
\end{abstract}



\keywords{Digital watermark, copyright protection, robust residual}



\maketitle

\section{Introduction}
Deep learning-based digital image watermarking techniques have made significant progress in recent years ~\cite{hosny2024digital,hu2024learning,wang2024must,jang2024waterf,fu2024waverecovery}. They enable the embedding of watermarks into arbitrary images in an imperceptible way, achieving tasks such as copyright protection and traceability. With the rapid development of large models, the need for copyright protection and traceability of generated content has become particularly critical ~\cite{dathathri2024scalable,zhang2024editguard,wang2023security,lyu2023pathway,wang2024lampmark,wu2023sepmark,fernandez2023stable,wang2024robust}. In response, various countries have issued corresponding standards requiring generative content to include identification mechanisms \cite{golda2024privacy,ye2024privacy,wang2024security}. However, current deep learning-based digital image watermarking techniques still face challenges in simultaneously optimizing visual quality, robustness, and generalization. Consequently, advancing deep learning watermarking technologies has become an urgent task.
\begin{figure}[tb]%[!t]
	\centering
	\includegraphics[trim = {90mm 33mm 75mm 33mm},clip,width=3.3in]{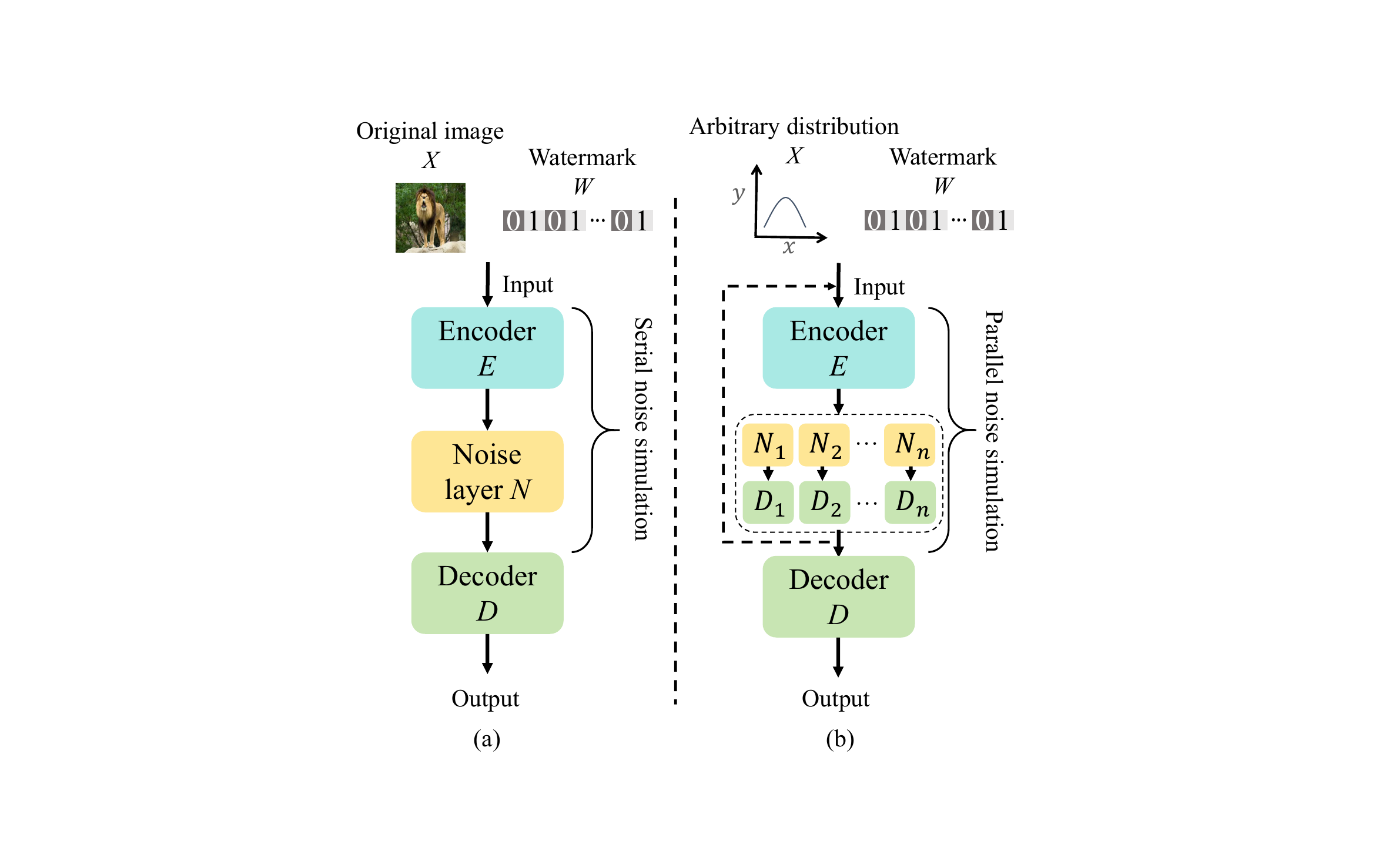}
	\caption{(a) Existing deep learning watermarking framework improves the robustness by employing serial noise simulation within the noise layer $N$. (b) Proposed framework in this work enhances the robustness by employing parallel noise simulation within the encoder $E$.}
	\label{fig1}
\end{figure}

In general, training a deep learning watermarking framework is a process of adversarial optimization. On the one hand, the network is trained to generate high-visibility watermarked images, ensuring that the watermarked images are visually indistinguishable from the original ones. However, the network must resist various noise attacks, such as JPEG compression, scaling, and Gaussian noise \cite{wan2022comprehensive,ma2023robust}. Achieving high visual quality can be done by modifying fewer pixels (with lower intensity) or embedding less watermark information. In contrast, enhancing robustness typically requires modifying more pixels (with higher intensity) or embedding more watermark information. During the optimization process, the critical challenge lies in designing the network architecture and loss function to balance visual quality and robustness. Currently, most deep learning watermarking frameworks adopt an encoder-noise layer-decoder (E-N-D) structure ~\cite{Zhu2018HiDDeNHD,tancik2020stegastamp,jia2021mbrs,ma2022towards,huang2023arwgan,9956019,ma2025geometric,10891427}, as shown in Figure \ref{fig1} (a). The encoder embeds the watermark into the original image to generate the watermarked image while ensuring its imperceptibility. The noise layer enhances the robustness by simulating the distortion process of watermarked images, enabling resistance to various noise attacks. The decoder extracts the watermark from the attacked watermarked images. In this framework, joint training with aggregated loss functions is typically used to optimize both visual quality and robustness \cite{jia2021mbrs,ge2023robust,10098654,10175655}. Additionally, to address the issue of gradient truncation caused by rounding operations in the noise layer, a two-stage training strategy can be used: first, optimizing the encoder parameters and freezing them, and then optimizing the decoder parameters to enable effective gradient backpropagation ~\cite{liu2019novel,yin2023anti,ma2025ropasst}. Although existing deep learning watermarking methods have addressed many underlying challenges, limitations persist in visual quality and robustness. Robustness largely depends on the design of the noise layer, which in most networks simply concatenates multiple noises to simulate composite attacks ~\cite{ge2023robust,huang2023arwgan,hosny2024digital,hu2024learning}. However, this approach struggles to reflect the characteristics of specific individual noise types and is insufficient for adapting to complex real-world scenarios. Meanwhile, robustness optimization often compromises visual quality, making it difficult to simultaneously optimize multiple objectives in the trained network. In summary, establishing an end-to-end deep learning watermarking framework remains an urgent and unresolved challenge.

In this work, we propose an adaptive robust iterative watermarking framework (ARIW-Framework) that addresses the multi-objective optimization challenges. Specifically, our network learns to generate a robust residual, which is added to the original image to produce the watermarked image. The resulting watermarked image is visually imperceptible from the original image. To ensure high visual quality, we use the gradient of the original image to determine the embedding strength at each pixel location. This ensures that more watermark is embedded in complex regions while less is embedded in smooth regions. To enhance robustness against various types of noise attacks, we adopt a parallel design within the encoder, concatenating multiple noise types to achieve robustness optimization for individual noise attacks, as shown in Figure \ref{fig1} (b). This design enables the encoder to generate watermarked images with both high visual quality and robustness. Additionally, our framework focuses solely on optimizing the robust residual, independent of the original image. In other words, our framework does not impose restrictions on the specific optimization target, any objective can be optimized to approach the domain of the optimal robust residual. In summary, the contributions can be summarized as bellow:
\begin{itemize}
	\item We propose an adaptive robust iterative watermarking framework capable of sampling any spatial distribution as the iterative target, without being constrained by the original image.
	\item We design an encoder with linearly additive residuals, which simulates distortions caused by various attack types in parallel, enabling the training of highly robust residuals.
	\item We introduce the use of original image gradients to determine the adaptive embedding strength at each pixel, enabling watermark embedding in complex regions while avoiding smooth regions.
	\item Experimental results demonstrate that our method achieves outstanding robustness against signal processing and geometric attacks while maintaining high visual quality of the watermarked images.
\end{itemize}

\begin{figure*}[htb]%[!t]
	\centering
	\includegraphics[trim = {0mm 33mm 0mm 33mm},clip,width=\linewidth]{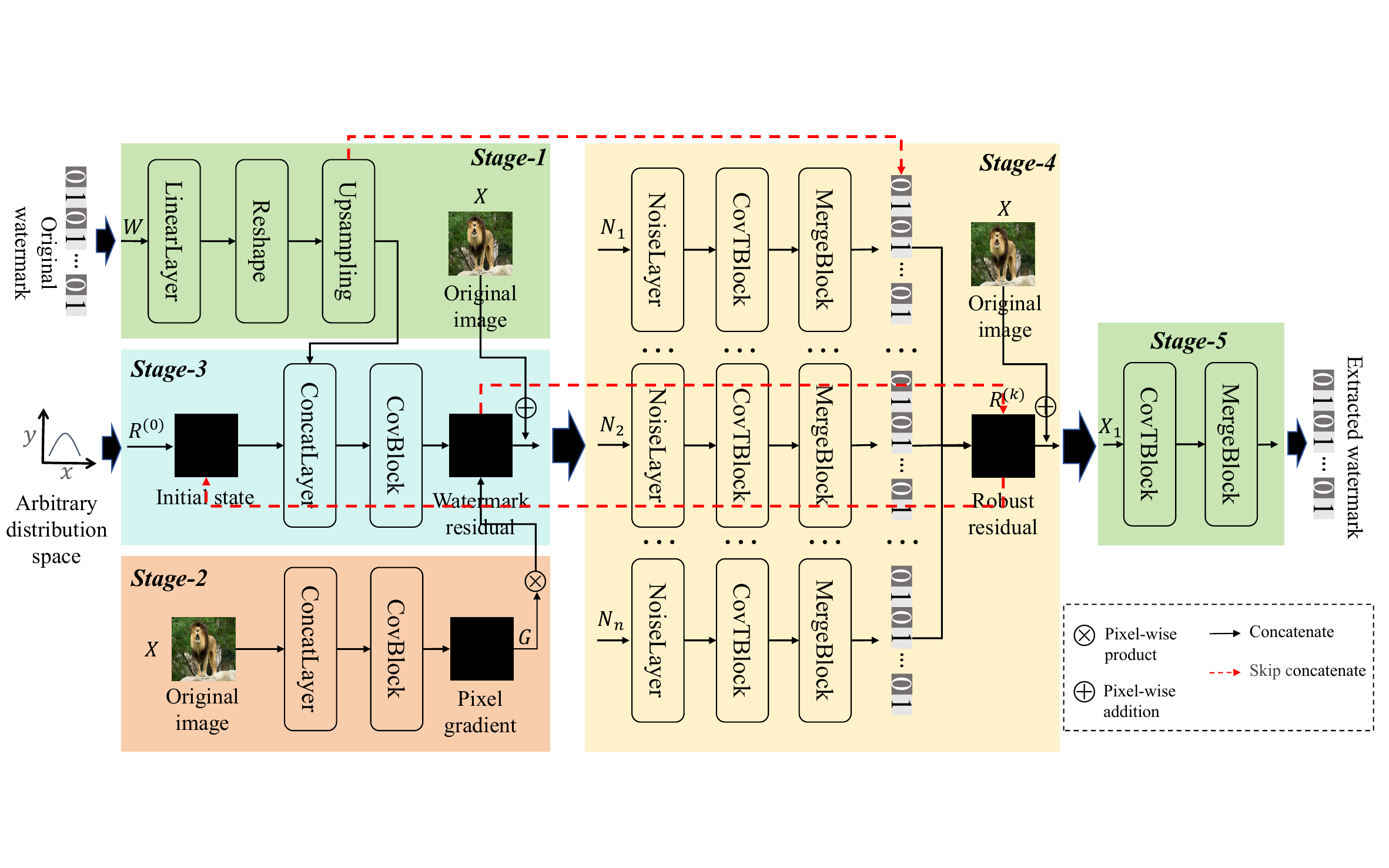}
	\caption{The pipeline of our framework.}
	\label{fig2}
\end{figure*}
\section{Method}
\subsection{Pipeline Overview}
%Our proposed method primarily focuses on scenarios involving copyright protection and provenance tracking of original image content or content generated by large models. By leveraging a trained network, we enable watermark embedding and extraction for images, with the overall process illustrated in Figure 1. To train a watermarking framework that balances visual quality and robustness, we introduce an adaptive robust iterative watermarking framework comprising five stages: Stage-1 preprocessing of the original watermark, Stage-2 computation of the original image gradient, Stage-3 iterative initialization (i.e., the process of finding the optimal residual), Stage-4 computation of robust residual weights (i.e., determining the robust residual weight for each attack type), and Stage-5 watermark extraction. Among these, stage Stage-2 ensures adaptive watermark embedding, while stages Stage-3 and Stage-4 are responsible for generating the watermark and robust residuals. Unlike existing encoder-noise-decoder (E-N-D) watermarking methods, our framework integrates the noise layer $N$ directly into the encoder to generate robust residuals. These residuals are then added to the original image to produce the watermarked image, which is subsequently passed to the decoder for watermark extraction, eliminating the need for separate distortion simulations of the watermarked image. The following sections provide a detailed explanation of our approach.
Our proposed method is primarily designed for scenarios involving copyright protection and traceability of original image content or content generated by large models. The proposed watermarking framework is illustrated in Figure \ref{fig2}. This framework is divided into five stages: \textit{Stage-1} Preprocessing of the original watermark, \textit{Stage-2} Calculation of the original image gradient, \textit{Stage-3} Iteration from the initial state (i.e., the process of finding the optimal residual), \textit{Stage-4} Calculation of robust residual weights (i.e., determining robust residual weights for each type of attack), \textit{Stage-5} Watermark extraction.
Unlike existing watermarking methods, the proposed framework integrates the noise layer $N$ directly into the encoder to generate highly robust residuals. The resulting residuals are added to the original image to produce the watermarked image, which is then passed to the decoder for watermark extraction without requiring additional distortion simulation on the watermarked image. The subsequent sections will provide a detailed explanation of our approach.

\subsection{Watermark Preprocessing}
The embedded watermark in this work is a sequence composed of 0s and 1s. Before feeding it into the network, a preprocessing step is required. This preprocessing primarily includes linear transformation and upsampling operations \cite{Tan_2024_CVPR}. Specifically, the linear transformation projects watermark vectors of fixed length into a higher-dimensional vector space to enhance their representational capacity. Meanwhile, the upsampling operation maps these vectors to higher-dimensional tensor spaces, enabling resolution adaptation to feature tensors of arbitrary sizes while simultaneously strengthening watermark representation capability to some extent. Let the original watermark be $W\in\{0,1\}^L$ and the original image be $X\in\mathbb{R}^{m\times n\times c}$. To enable successful upsampling of the watermark, $W$ needs to undergo a linear transformation, mapping it to a feature 
$W_1\in\mathbb{R}^{L_1}$:
\begin{equation}
	\label{eq1}
	f:W\in\{0,1\}^L\rightarrow W_1\in\{0,1\}^{L_1}
\end{equation}
subsequently, through reshape and upsampling operations, $W_1$ is mapped to a spatial representation $W_2\in\mathbb{R}^{m\times n\times c}$:
\begin{equation}
	\label{eq2}
	f:W_1\in\mathbb{R}^{L_1}\rightarrow W_2\in\mathbb{R}^{m\times n\times c}
\end{equation}
here, the upsampling factor is $d=m\times n\times c/L_1$. The resulting $W_2$ has the same resolution as the original image $X$, facilitating seamless concatenation and establishing the relationship between the watermark and the feature map. %Notably, the upsampling operation inherently induces substantial redundancy in the watermark payload. Specifically, the model ultimately embeds multiple replicated watermark instances rather than a singular watermark. This architectural design yields two principal advantages: (i) it facilitates convergence of the watermark cross-entropy loss during model optimization, and (ii) it enhances the robustness and accuracy of watermark extraction.

\subsection{Gradient Calculation}
Gradient computation is primarily used to determine the watermark embedding strength for each pixel. Since modifications in smooth regions of an image are more likely to cause noticeable visual changes, the watermark embedding should focus on complex regions while avoiding smooth areas to ensure high visual quality of the watermarked image. We use the gradient magnitude at each pixel position to measure the complexity of the region: the larger the absolute gradient value, the higher the texture complexity, and vice versa. Denoting the image gradient as $G\in\mathbb{R}^{m\times n\times c}$, then after the encoder 
$E$ generates the watermark residual $R$, the adaptive watermark residual can be expressed as:
\begin{equation}
	\label{eq3}
	R=G*R
\end{equation}

To efficiently compute the image gradient, we leverage TensorFlow's built-in automatic differentiation mechanism \cite{griewank2008evaluating,baydin2018automatic}. Using the encoder $E$ as the computational process, we record the forward propagation and then utilize backpropagation through the computational tape to obtain the gradient values. The principles of this calculation can be found in the documentation at https://github.com/tensorflow.

\subsection{Residual Iterative Optimization}\label{sec_RIO}
This section introduces the proposed robust watermark iterative framework, detailing the design of the encoder, robust weight calculation, decoder, and loss functions. For clarity, we denote the encoder, noise layer, and decoder as 
$E$, $N$ and $D$, respectively. Ideally, the watermarked image $X_1$ generated by the encoder $E$, is related to the original image $X$ by a residual $R$, expressed as:
\begin{equation}
	\label{eq4}
	X_1=E(X)=X+R
\end{equation}
The fundamental problem is to find a function $E$ such that the space of the original image $X$ can be mapped to the space of the watermarked image $X_1$:
\begin{equation}
	\label{eq5}
	\begin{split}
		E&:X\in\mathbb{R}^{m\times n\times c}\rightarrow X_1\in\mathbb{R}^{m\times n\times c}
	\end{split}
\end{equation}
such that
\begin{equation}
	\label{eq6}
	\begin{split}
		\phi_i(X,X_1)>\beta_i
	\end{split}
\end{equation}
where $\phi_i$ is a conditional restriction function, such as watermark accuracy, average peak signal-to-noise ratio (PSNR), structural similarity (SSIM) \cite{wang2004image}, etc., and $\beta_i$ is a real number.
This problem is equivalent to finding a function $E$ that maps the zero space to the residual space $R$:
\begin{equation}
	\label{eq7}
	E:0\in\mathbb{R}^{m\times n\times c}\rightarrow X_1-X=R\in\mathbb{R}^{m\times n\times c}
\end{equation}
the task is to enable the encoder $E$ to determine the optimal residual $R$ such that $\phi_i(X,X+R)>\beta_i$. Recognizing that Equation (\ref{eq7}) maps from the zero space to the residual space $R$, a natural question arises: can it map from any arbitrary distribution space to the residual space? The answer is affirmative. Transforming Equation (\ref{eq7}):
\begin{equation}
	\label{eq8}
	E:0+\theta\in\mathbb{R}^{m\times n\times c}\rightarrow R+\theta\in\mathbb{R}^{m\times n\times c}
\end{equation}
suggests that finding $R$ involves an intermediate distribution $\theta$, which is further optimized into the residual space $R$. To find the optimal residual 
$R$ such that $X_1=X+R$ satisfies the constraints $\phi_i(X,X+R)>\beta_i$, we propose the following iterative optimization format:
\begin{equation}
	\label{eq9}
	R^{(k+1)} = \alpha*\sum_i^{N_n}\omega_iR^{(k)},k=1,2,3,...
\end{equation}
where 
\begin{equation}
	\label{eq10}
	\begin{split}
		R^{(k)} &= G*E(R^{(k-1)},W),k=1,2,3,...\\
	\end{split}
\end{equation}
$\omega_i$ represents the residual weights of the i-th attack type, $\alpha$ denotes the embedding strength, and $G$ is the gradient. The extracted watermark $W^\prime$ is given by 
\begin{equation}
	\label{eq11}
	\begin{split}
		W^\prime=D(N(X+R^{(k)}))
	\end{split}
\end{equation}

Using Equation (\ref{eq9}), starting from an initial $R^{(0)}$, the iterative method enables finding the optimal solution. Based on Equation (\ref{eq9}), we design an encoder $E$, noise layer $N$, and decoder $D$ to guide the network in finding the optimal residual $R$ that satisfies the constraints. The following sections will provide a detailed explanation of these components.

\subsubsection{Encoder}
The encoder $E$ serves as a mapping from an arbitrary space to the residual space, primarily for generating watermark residuals and robust residuals. To achieve high visual quality for the watermarked images, the watermark residual should be as minimal as possible. Simultaneously, to enhance the robustness of the watermark residual, it is necessary to simulate image distortions caused by various attack types and refine the pixel values of the residuals, thereby obtaining robust residuals. To generate watermark residuals, we use multiple convolutional layers \cite{he2016deep,he2016identity} to extract residual features and establish relationships between each feature map and the watermark $W_2$. This design facilitates the localization of watermark positions during decoding and enables fast extraction of watermark information. Specifically, for each feature map $x_i$ in the encoder:
\begin{equation}
	\label{eq12}
	\begin{split}
		f_E:x_i\in\mathbb{R}^{m_i\times n_i\times c_i}\rightarrow x^\prime_i\in\mathbb{R}^{m_i\times n_i\times c_i+3}
	\end{split}
\end{equation}
where $m_i\times n_i\times c_i$ is the resolution of the feature map $x_i$ of the current layer $i$.

To ensure the robustness of the watermark residuals, we incorporate robust weight calculations for various attack types into the encoder, as shown in the \textit{Stage-4} of Figure \ref{fig2}. This process involves simulating image distortions caused by each attack type and extracting watermark information, enabling the computation of the cross-entropy loss between the extracted watermark and the original watermark. This loss serves as the robust weight for the current attack type and is also treated as a local loss for joint optimization within the overall framework. Specifically, for each attack type, the distortion simulation is as follows:
\begin{equation}
	\label{eq13}
	\begin{split}
		f_N:X+R^{(k)}\in\mathbb{R}^{m\times n\times c}\rightarrow X^\prime\in\mathbb{R}^{m\times n\times c}
	\end{split}
\end{equation}
where $X^\prime$ is the watermarked image subjected to attack type $N_i$. The attacked image $X^\prime$ is fed into the decoder to extract the watermark $W^\prime=D(X^\prime)$(the structure of 
$D$ will be described later). The robust weight for the current attack type $N_i$ then calculated based on this extraction:
\begin{equation}
	\label{eq14}
	\omega_i=[softmax(\omega)]_i=\dfrac{\exp(\omega_i)}{\sum_j\exp(\omega_j)}=\dfrac{\exp(Acc_i)}{\sum_j\exp(Acc_j)}
\end{equation}
where $Acc_j$ is the accuracy or cross entropy of the original watermark $W$ and the extracted watermark $W^\prime$.

\subsubsection{Decoder}
The decoder is responsible for extracting the watermark from the watermarked image. To facilitate rapid localization of the watermark positions and efficient extraction of watermark information, the decoder employs a multi-layer deconvolution \cite{noh2015learning} structure, with the number of channels symmetrically aligned with the encoder. This design ensures that the decoder can effectively return to the feature space of the encoder during feature extraction, enabling the retrieval of the watermark information embedded in the channels. Specifically, for each feature map $x_i$ in the decoder:
\begin{equation}
	\label{eq12}
	\begin{split}
		f_D:x_i\in\mathbb{R}^{m_i\times n_i\times c_i}\rightarrow W^\prime_i\in\mathbb{R}^{m_i\times n_i\times 3}
	\end{split}
\end{equation}
where $W^\prime_i$ represents the watermark information decoupled from the current feature map $x_i$. 

Since each feature map in decoder $D$ can decouple watermark information, our framework incorporates an aggregation layer at the final decoder stage to fuse these watermark components, thereby significantly enhancing robustness. The aggregation layer employs dual operations: (1) channel-wise summation and (2) channel-wise multiplication:
\begin{equation}
	\label{eq15}
	W^{\prime}_{sum} = \sum^l_{i=1}W^\prime_i \quad \text{and} \quad W^\prime_{prod} = \prod_{i=1}^lW^\prime_i
\end{equation}
where $l$ denotes the number of convolutional layers in the decoder $D$. The aggregated features $W^{\prime}_{sum}$ and $W^\prime_{prod}$ are then concatenated and fed into a final dense layer (whose output dimension matches the original watermark size), followed by a Sigmoid activation function to produce the extracted watermark. This aggregation layer effectively suppresses anomalous watermark components, ensuring stable and robust watermark generation.

%Since watermark information can be decoupled from each layer's feature map, the decoder incorporates an aggregation layer at its final stage to merge these extracted watermark details, thereby further enhancing robustness. The merge layer employs two methods—channel summation and channel multiplication. These operations effectively suppress anomalous watermark information from specific locations, ensuring that the final extracted watermark output is stable and reliable.

\subsubsection{Loss Function}
The proposed method employs an end-to-end joint training approach to simultaneously optimize the encoder and decoder. The loss function primarily consists of two components: image loss and watermark loss.

Image loss includes mean squared error (MSE) \cite{goodfellow2016deep} loss $\mathcal{L}_1$ and PSNR \cite{wang2004image} loss $\mathcal{L}_2$, both of which measure the visual quality between the original image $X$ and the watermarked image 
$X^\prime$. The $\mathcal{L}_1$ is expressed as:
\begin{equation}\label{loss_1}
	\begin{split}
		\mathcal{L}_1&=\frac{1}{m\times n}\sum_{i=0}^{m-1}\sum_{j=0}^{n-1}(X^\prime(i,j)-X(i,j))^2\\
		&=\frac{1}{m\times n}\sum_{i=0}^{m-1}\sum_{j=0}^{n-1}R^2
	\end{split}
\end{equation}
the $\mathcal{L}_2$ is expressed as:
\begin{equation}\label{loss_2}
	\mathcal{L}_2=\dfrac{1}{PSNR(X,X^\prime))}
\end{equation}

Watermark loss is optimized using cross-entropy \cite{mao2023cross} and includes global and local watermark losses. The global watermark loss measures the cross-entropy between the original watermark $W$ and the final extracted watermark $W^\prime$:
\begin{equation}\label{loss_3}
	\mathcal{L}_3=\frac{1}{L}\sum_{j=0}^{L-1}-w_{j}\log(w^{\prime}_{j})-(1-w_{j})\log(1-w^{\prime}_{j}) 
\end{equation}
where $L$ is the length of the watermark $W$, $w^{\prime}_j$ represents the watermark extracted by the decoder $D$ at the j-th position, and its value is 0 or 1.

The local watermark loss evaluates the cross-entropy between the original watermark $W$ and the extracted watermark $W^\prime_{N_i}$ during robust weight calculation: 
\begin{equation}\label{loss_3}
	\begin{split}
		\mathcal{L}_4&=\sum_{i=1}^n\mathcal{L}_{N_i}\\
		&=\frac{1}{L}\sum_{i=1}^n\sum_{j=0}^{L-1}-w_{i,j}\log(w^{\prime}_{i,j})-(1-w_{i,j})\log(1-w^{\prime}_{i,j}) 
	\end{split}
\end{equation}
where $w^{\prime}_{i,j}$ represents the watermark bit extracted at the j-th position under the i-th type noise $N_i$, whose value is 0 or 1. $\mathcal{L}_{N_i}$ serves as the weight $\omega_i$ of each robust residual $R_{N_i}$, which is dynamically optimized with the number of iterations.

Therefore, the final loss function of the network is:
\begin{equation}\label{loss_5}
	\mathcal{L}_{Total}=\lambda_1\mathcal{L}_{1}+\lambda_2\mathcal{L}_{2}+\lambda_3\mathcal{L}_{3}+\lambda_4\mathcal{L}_{4}
\end{equation}
where $\lambda_1$, $\lambda_2$, $\lambda_3$ and $\lambda_4$ are the corresponding loss weights, whose initial values are $(1.5,1.0,1.0,1.0)$. These weights can be adjusted during training to balance the contributions of each loss component.

\section{Experiment}
To validate the proposed method's performance in terms of visual quality, robustness, and generalization, this section conducts experimental evaluations across multiple datasets and various noise attacks. Additionally, ablation studies are performed to further demonstrate the rationality of the proposed method, while comparative experiments highlight its advantages.

\subsection{Experimental Setup}
\subsubsection{Dataset}
The datasets include BOSSBase \cite{bas2011break}, Mirflickr \cite{huiskes2008mir}, and COCO \cite{lin2014microsoft}. For the training set, we randomly selected 2,000 color images from the Mirflickr database. For the test set, we randomly selected 100 images from each of the BOSSBase (grayscale images), Mirflickr (color images), and COCO (color images) databases. Additionally, since the proposed method requires input images to have a fixed size of 400×400, all images are resized to this resolution prior to the experiments.

\subsubsection{Evaluation Metric}
The evaluation methods primarily focus on the visual quality and robustness of the generated watermarked images. For visual quality, we use the PSNR and SSIM \cite{wang2004image} as metrics, where higher values indicate better visual quality of the watermarked images. For robustness, we measure the average bit accuracy, which evaluates the accuracy of watermark extraction under various types of attacks. Additionally, tests are conducted on multiple datasets to validate the generalization performance of the proposed method.

\subsubsection{Implementation Detail}
The experiments are conducted on a platform running the Windows 10 operating system, equipped with an Intel(R) Xeon(R) Gold 6161 CPU @ 2.20GHz and 128 GB of memory. The implementation is carried out using Python, with Python 3.6 as the programming and compilation environment. For training, the hyperparameter settings are as follows: the optimizer used is Adam, with a learning rate of 0.0001, a batch size of 1, and 140,000 iterations. The length of the watermark information is set to 100, the convolution kernel size is 3×3, the stride is 1, and the embedding strength is 1.0.

\begin{table}[h]
	\centering
	\caption{The PSNR and SSIM of three test sets at different embedding strengths.}
	\label{tab:booktabs1}
	\resizebox{\linewidth}{!}{\begin{tabular}{lrrrrrrrrrrr}
			\toprule
			Dataset & Metric & 0.2 & 0.4 & 0.6 & 0.8 & 1.0 & 1.2 & 1.4 & 1.6 & 1.8 & 2.0 \\
			\midrule
			Mirflickr & PSNR$\uparrow$ &50.478 &48.140 &45.860 &43.791 &42.084 &40.753 &39.494 &38.483 &37.555 &36.708\\
			& SSIM$\uparrow$ &0.998 &0.996 &0.994 &0.991 &0.987 &0.983 &0.979 &0.974 &0.969 &0.964\\
			\midrule
			COCO & PSNR$\uparrow$ &50.255 &47.815 &45.451 &43.433 &41.762 &40.326 &39.102 &38.064 &37.158 &36.324\\
			& SSIM$\uparrow$ &0.998 &0.997 &0.994 &0.992 &0.988 &0.984 &0.980 &0.976 &0.971 &0.966\\
			\midrule
			BOSSBase & PSNR$\uparrow$ &50.306 &47.975 &45.656 &43.666 &42.023 &40.594 &39.403 &38.392 &37.432 &36.580\\
			& SSIM$\uparrow$ &0.998 &0.996 &0.994 &0.991 &0.987 &0.983 &0.979 &0.974 &0.969 &0.964\\
			\bottomrule
	\end{tabular}}
	
\end{table}

\begin{table}[h]
	\centering
	\caption{The robustness of three test sets at different embedding strengths (\%).}
	\label{tab:booktabs2}
	\resizebox{\linewidth}{!}{\begin{tabular}{lrrrrrrrrr}
			\toprule
			Dataset & $\alpha$ & Identity & \makecell{JPEG\\(QF=50)} & \makecell{Gaussian\\noise\\($\sigma$=0.02)} & \makecell{Gaussian\\filter\\($k$=7)} & \makecell{Crop\\($p$=0.03)} & \makecell{Cropout\\($p$=0.9)} & \makecell{Dropout\\($p$=0.9)} & \makecell{Scaling\\($p$=0.5)}\\
			\midrule
			Mirflickr & 1.0 &100.0 &99.66 &99.96 &99.99 &99.71 &99.98 &99.99 &99.94 \\
			& 0.8 &99.98 &99.16 &99.93 &99.96 &99.81 &99.98 &100.0 &99.82 \\
			& 0.6 &99.95 &97.13 &99.71 &99.98 &99.12 &99.93 &99.97 &99.33 \\
			& 0.4 &99.83 &91.39 &98.81 &99.73 &97.70 &99.84 &99.77 &97.63 \\
			& 0.2 &97.87 &75.78 &91.17 &97.87 &93.29 &97.93 &97.69 &91.44 \\
			\midrule
			COCO & 1.0 &100.0 &99.54 &99.98 &99.98 &99.88 &99.98 &99.99 &99.94 \\
			& 0.8 &99.98 &98.87 &99.94 &99.98 &99.57 &100.0 &99.99 &99.76 \\
			& 0.6 &99.88 &96.51 &99.67 &99.88 &98.74 &99.91 &99.91 &99.20 \\
			& 0.4 &99.49 &90.79 &98.32 &99.39 &97.42 &99.35 &99.42 &97.21 \\
			& 0.2 &96.39 &76.68 &90.37 &96.09 &91.25 &96.08 &95.99 &89.38 \\
			\midrule
			BOSSBase & 1.0 &100.0 &99.80 &99.99 &100.0 &99.97 &100.0 &100.0 &100.0 \\
			& 0.8 &100.0 &99.32 &99.99 &100.0 &99.90 &100.0 &100.0 &99.98 \\
			& 0.6 &99.99 &97.32 &99.91 &100.0 &99.80 &100.0 &100.0 &99.81 \\
			& 0.4 &99.93 &91.59 &99.19 &99.94 &98.85 &99.90 &99.91 &99.19 \\
			& 0.2 &98.87 &77.20 &93.52 &98.91 &95.71 &98.94 &98.88 &95.53 \\
			\bottomrule
	\end{tabular}}
	
\end{table}

\begin{figure*}[!htb]%[!t]
	\centering
	\includegraphics[trim = {15mm 20mm 15mm 20mm},clip,width=\linewidth]{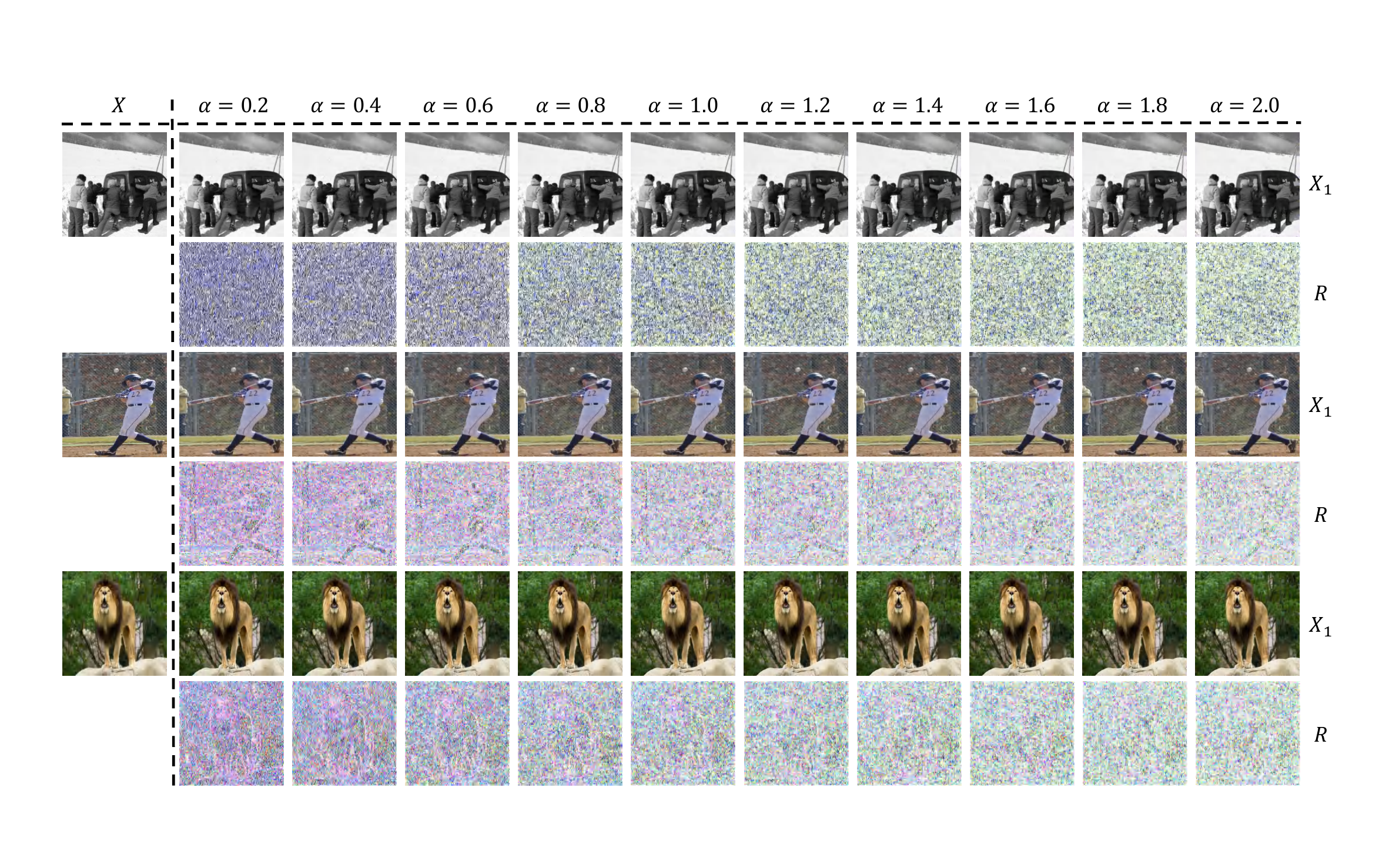}
	\caption{Watermarked image $X_1$ and residual image $R$ generated under different embedding strengths.}
	\label{fig3}
\end{figure*}

\begin{table*}[!h]
	\centering
	\caption{The visual quality and robustness of different watermarking methods (\%).}
	\label{tab:booktabs3}
	\resizebox{\linewidth}{!}{\begin{tabular}{c c c c c c c c c c c c c c c}
			%\hline
			\toprule
			Dataset &Method &$m\times n$ &$L$&PSNR$\uparrow$&SSIM$\uparrow$& Identity & \makecell{JPEG\\(QF=50)} & \makecell{Gaussian\\noise\\($\sigma$=0.02)} & \makecell{Gaussian\\filter\\($k$=7)} & \makecell{Crop\\($p$=0.03)} & \makecell{Cropout\\($p$=0.9)} & \makecell{Dropout\\($p$=0.9)} & \makecell{Scaling\\($p$=0.5)}&Average\\
			\midrule
			\multirow{9}{*}{Mirflickr}&HiDDeN\cite{Zhu2018HiDDeNHD}&$128\times128$&30&32.5588&0.9373&93.75&91.83&97.76&88.03&75.37&91.53&97.50&78.23&89.25\\
			&StegaStamp\cite{tancik2020stegastamp}&$400\times400$&100&29.302&0.890&99.93&99.89&99.84&99.92&99.61&99.32&99.85&99.84&99.78\\
			&MBRS\cite{jia2021mbrs}&$256\times256$&256&41.99&0.988&100.0&99.36&81.12&55.47&49.78&94.22&100.0&49.93&78.74\\
			&CIN\cite{ma2022towards}&$128\times128$&30&41.437&0.980&100.0&99.67&100.0&100.0&99.93&100.0&98.30&98.93&99.60\\
			&ARWGAN\cite{huang2023arwgan}&$128\times128$&30&22.647&0.877&99.16&90.27&94.50&99.13&95.80&98.87&99.23&95.83&96.60\\
			&Document\cite{ge2023robust}&$400\times400$&100&32.133&0.991&99.13&99.16&98.83&95.45&52.85&99.22&99.21&98.53&92.80\\
			&De-END\cite{9956019}&$128\times128$&64&48.750&0.951&98.48&70.41&91.34&99.90&--&97.56&100.0&--&92.95\\
			&ST-DCN$^\star$\cite{ma2025geometric}&$224\times224$&196&35.973&0.992&100.0&99.93(40)&98.09(0.04)&100.0&96.75(0.2)&96.87(0.5)&95.44(0.7)&99.38(0.4)&98.31\\
			&\textbf{\underline{Ours}}&$400\times400$&100&42.084&0.987&100.0 &99.66 &99.96 &99.99 &99.71 &99.98 &99.99 &99.94&\textbf{\underline{99.90}}\\
			\midrule
			\multirow{9}{*}{COCO}&HiDDeN\cite{Zhu2018HiDDeNHD}&$128\times128$&30&32.893&0.947&98.33&92.73&97.93&89.93&74.22&92.23&98.23&79.13&90.34\\
			&StegaStamp\cite{tancik2020stegastamp}&$400\times400$&100&30.068&0.910&99.88&99.82&99.88&99.88&99.62&99.18&99.79&99.90&99.74\\
			&MBRS\cite{jia2021mbrs}&$256\times256$&256&42.008&0.988&100.0&99.62&79.57&100.0&49.61&93.96&100.0&50.36&84.14\\
			&CIN\cite{ma2022towards}&$128\times128$&30&41.909&0.983&100.0&99.77&100.0&100.0&99.97&100.0&97.40&99.27&99.55\\
			&ARWGAN\cite{huang2023arwgan}&$128\times128$&30&31.736&0.959&99.20&94.86&96.13&99.20&96.23&99.23&99.06&96.30&97.53\\
			&Document\cite{ge2023robust}&$400\times400$&100&32.299&0.993&99.96&99.95&99.44&99.89&53.19&99.42&99.51&99.42&93.85\\
			&De-END\cite{9956019}&$128\times128$&64&49.013&0.990&98.41&70.36&96.58&99.98&--&97.75&100.0&--&93.85\\
			&ST-DCN$^\star$\cite{ma2025geometric}&$224\times224$&196&35.973&0.992&100.0&99.93(40)&98.09(0.04)&100.0&96.75(0.2)&96.87(0.5)&95.44(0.7)&99.38(0.4)&98.31\\
			&\textbf{\underline{Ours}}&$400\times400$&100&41.762&0.988&100.0 &99.54 &99.98 &99.98 &99.88 &99.98 &99.99 &99.94&\textbf{\underline{99.91}}\\
			\midrule
			\multirow{9}{*}{BOSSBase}&HiDDeN\cite{Zhu2018HiDDeNHD}&$128\times128$&30&35.940&0.970&97.00&87.27&98.76&88.77&75.70&92.77&98.47&77.53&89.53\\
			&StegaStamp\cite{tancik2020stegastamp}&$400\times400$&100&30.068&0.910&99.99&99.97&99.92&99.97&99.75&99.60&99.93&99.91&99.88\\
			&MBRS\cite{jia2021mbrs}&$256\times256$&256&42.975&0.988&100.0&99.87&78.99&99.30&49.24&94.42&100.0&50.18&84.00\\
			&CIN\cite{ma2022towards}&$128\times128$&30&43.219&0.984&100.0&99.60&100.0&100.0&100.0&100.0&96.40&99.30&99.41\\
			&ARWGAN\cite{huang2023arwgan}&$128\times128$&30&31.736&0.959&95.87&85.93&93.09&99.30&97.20&99.17&98.993&97.70&95.91\\
			&Document\cite{ge2023robust}&$400\times400$&100&32.299&0.993&99.96&99.98&99.95&99.89&53.19&99.97&99.99&99.96&94.11\\
			&De-END\cite{9956019}&$128\times128$&64&49.126&0.981&100.0&65.18&98.37&100.0&--&98.32&100.0&--&93.65\\
			&ST-DCN$^\star$\cite{ma2025geometric}&$224\times224$&196&35.973&0.992&100.0&99.93(40)&98.09(0.04)&100.0&96.75(0.2)&96.87(0.5)&95.44(0.7)&99.38(0.4)&98.31\\
			&\textbf{\underline{Ours}}&$400\times400$&100&42.023&0.987&100.0 &99.80 &99.99 &100.0 &99.97 &100.0 &100.0 &100.0&\textbf{\underline{99.97}}\\
			\bottomrule
	\end{tabular}}
	
\end{table*}

\begin{table*}[!h]
	\centering
	\caption{The visual quality and robustness corresponding to different ablation experiments (\%).}
	\label{tab:booktabs4}
	\resizebox{\linewidth}{!}{\begin{tabular}{lrrrrrrrrrrrr}
			\toprule
			Ablation module & Dataset & PSNR$\uparrow$ & SSIM$\uparrow$ & Identity & \makecell{JPEG\\(QF=50)} & \makecell{Gaussian\\noise\\($\sigma$=0.02)} & \makecell{Gaussian\\filter\\($k$=7)} & \makecell{Crop\\($p$=0.03)} & \makecell{Cropout\\($p$=0.9)} & \makecell{Dropout\\($p$=0.9)} & \makecell{Scaling\\($p$=0.5)}&Average\\
			\midrule
			$X^{(0)}$=1 & Mirflickr &42.084 &0.987 &100.0 &99.66 &99.96 &99.99 &99.711 &99.98 &99.99 &99.94 &\textbf{\underline{99.90}}\\
			& COCO &41.762 &0.988 &100.0 &99.54 &99.98 &99.98 &99.88 &99.98 &99.99 &99.94 &\textbf{\underline{99.91}}\\
			& BOSSBase &42.023 &0.987 &100.0 &99.80 &99.99 &100.0 &99.97 &100.0 &100.0 &100.0 &\textbf{\underline{99.97}}\\
			\midrule
			$X^{(0)}$=0 & Mirflickr &40.704 &0.991 &100.0 &98.64 &99.99 &100.0 &86.21 &100.0 &100.0 &99.84 &98.09\\
			& COCO &40.159 &0.992 &100.0 &99.09 &99.98 &99.98 &88.13 &99.99 &100.0 &99.77 &98.37\\
			& BOSSBase &40.158 &0.991 &100.0 &99.40 &100.0 &100.0 &93.48 &100.0 &100.0 &99.98 &99.11\\
			\midrule
			$X^{(0)}$=$X$ & Mirflickr &38.935 &0.984 &100.0 &99.85 &100.0 &100.0 &98.02 &100.0 &100.0 &99.98 &99.73\\
			& COCO &38.625 &0.986 &100.0 &99.85 &99.99 &100.0 &97.46 &100.0 &100.0 &99.98 &99.66\\
			& BOSSBase &38.858 &0.984 &100.0 &99.87 &100.0 &100.0 &99.28 &100.0 &100.0 &100.0 &99.89\\
			\midrule
			$X^{(0)}\sim N(0,1)$ & Mirflickr &38.170 &0.974 &100.0 &99.50 &99.99 &100.0 &99.59 &100.0 &100.0 &99.44 &99.82\\
			& COCO &37.788 &0.975 &100.0 &99.53 &99.98 &99.99 &99.59 &100.0 &100.0 &99.60 &99.84\\
			& BOSSBase &37.720 &0.970 &100.0 &99.76 &100.0 &100.0 &99.91 &100.0 &100.0 &99.93 &99.95\\
			\midrule
			Without $G$ & Mirflickr &37.812 &0.988 &100.0 &99.88 &99.99 &99.97 &91.13 &100.0 &100.0 &99.86 &98.85\\
			& COCO &37.395 &0.989 &100.0 &99.93 &100.0 &100.0 &90.09 &99.98 &100.0 &99.74 &98.72\\
			& BOSSBase &37.369 &0.988 &100.0 &99.96 &100.0 &100.0 &91.91 &100.0 &100.0 &99.97 &98.98\\
			\midrule
			$1\times1$ & Mirflickr &28.290 &0.932 &99.63 &98.80 &99.67 &99.77 &95.38 &99.60 &99.81 &99.19 &98.98\\
			& COCO &27.225 &0.940 &99.89 &99.42 &99.86 &99.90 &95.94 &99.87 &99.84 &99.57 &99.29\\
			& BOSSBase &27.706 &0.936 &99.98 &99.86 &99.94 &99.98 &97.47 &99.97 &99.97 &99.94 &99.64\\
			\midrule
			$5\times5$ & Mirflickr &41.257 &0.992 &100.0 &99.53 &100.0 &100.0 &96.74 &100.0 &100.0 &99.90 &99.52\\
			& COCO &40.697 &0.992 &100.0 &99.49 &99.98 &99.99 &97.15 &100.0 &100.0 &99.86 &99.56\\
			& BOSSBase &40.896 &0.991 &100.0 &99.81 &99.99 &100.0 &98.85 &100.0 &100.0 &99.98 &99.83\\
			\midrule
			$7\times7$ & Mirflickr &39.498 &0.990 &100.0 &99.52 &99.95 &100.0 &89.27 &99.99 &100.0 &99.86 &98.57\\
			& COCO &38.935 &0.991 &99.97 &99.43 &99.97 &99.99 &88.64 &99.98 &99.96 &99.82 &98.47\\
			& BOSSBase &39.283 &0.990 &100.0 &99.74 &99.98 &100.0 &91.10 &100.0 &100.0 &99.97 &98.85\\
			\midrule
			SA & Mirflickr &43.499 &0.992 &99.99 &93.33 &99.93 &99.99 &98.91 &99.99 &99.99 &99.37 &98.94\\
			& COCO &43.045 &0.993 &99.93 &94.02 &99.78 &99.88 &98.57 &99.94 &99.96 &99.24 &98.91\\
			& BOSSBase &43.168 &0.992 &100.0 &96.33 &99.98 &100.0 &99.63 &100.0 &100.0 &99.94 &99.49\\
			\midrule
			CA & Mirflickr &45.104 &0.993 &100.0 &94.08 &99.92 &100.0 &99.10 &100.0 &99.99 &99.86 &99.12\\
			& COCO &44.754 &0.994 &99.97 &92.84 &99.85 &99.93 &99.15 &99.96 &99.96 &99.79 &98.93\\
			& BOSSBase &44.863 &0.993 &99.99 &94.40 &99.99 &100.0 &99.59 &99.98 &100.0 &99.94 &99.24\\
			\midrule
			CBAM & Mirflickr &46.744 &0.996 &99.62 &77.88 &98.45 &99.64 &94.71 &99.59 &99.57 &98.11 &95.95\\
			& COCO &46.338 &0.996 &99.25 &79.00 &98.28 &99.38 &95.42 &99.24 &99.21 &97.85 &95.95\\
			& BOSSBase &46.513 &0.996 &99.93 &76.56 &99.61 &99.92 &98.84 &99.94 &99.91 &99.73 &96.80\\
			\bottomrule
	\end{tabular}}
\end{table*}

\subsection{Visual Quality}
This section presents the visual effects of watermarked images generated by the proposed method under varying embedding strengths. As shown in Table \ref{tab:booktabs1}, our method achieves high-quality visual results across multiple datasets, demonstrating strong generalization capabilities. 
Notably, at an embedding strength of $\alpha=1.0$, the visual quality achieves PSNR $>$41dB and SSIM$>$0.98. When increasing to $\alpha=2.0$, the metrics remain excellent with PSNR$>$36dB and SSIM$>$0.96. These results demonstrate that our model maintains superior image generation quality even under high embedding strengths, providing enhanced robustness for practical applications where perfect visual fidelity is not critical. Furthermore, as illustrated in Figure \ref{fig3}, the proposed method produces visually imperceptible modifications. This remarkable performance stems from our adaptive embedding strength strategy during network optimization, where image gradients dynamically regulate the embedding intensity to optimize both visual quality and watermark robustness.

%Specifically, when the embedding strength $\alpha=1.0$, the visual quality metrics reach a PSNR exceeding 41.0dB and an SSIM greater than 0.98. Furthermore, as illustrated in Figure 1, the modifications introduced by our method are imperceptible to the human eye, indicating that the proposed approach effectively generates minimal robust residuals $R$. These residuals, when added to the original image $X$, produce the watermarked image $X_1$ with high imperceptibility, ensuring the watermark remains visually undetectable.

\subsection{Robustness}
To evaluate robustness, we tested the watermark extraction accuracy under various noise attacks, including Identity (no distortion), JPEG compression (quality factor QF=50), Gaussian noise (variance $\sigma=0.02$), Gaussian filtering (variance $\sigma=0.02$, kernel size $k=7$), Dropout (rate $p=0.9$), Cropout (rate $p=0.9$), Crop (rate $p=0.03$), and Scaling (scaling factor 
$p=0.5$) \cite{9956019,ma2025geometric}. To verify that the proposed method can produce high-visual-quality watermarked images while maintaining excellent robustness, we primarily assessed robustness at embedding strengths $\alpha\leq1.0$.

As shown in Table \ref{tab:booktabs1} and Table \ref{tab:booktabs2}, the proposed method demonstrates strong robustness across different types of attacks. When the embedding strength $\alpha=1.0$, the accuracy under all noise attacks exceeds 99\%, with a visual quality of PSNR over 41dB and SSIM above 0.98. Even at a lower embedding strength of $\alpha=0.4$, the method achieves over 90\% accuracy under all noise attacks, with a PSNR exceeding 47dB and SSIM above 0.99. These results indicate that our method maintains excellent robustness while generating high-visual-quality watermarked images, offering a flexible range of generation schemes to meet diverse practical application requirements.

\begin{figure*}[!htb]%[!t]
	\centering
	\includegraphics[trim = {0mm 40mm 0mm 50mm},clip,width=\linewidth]{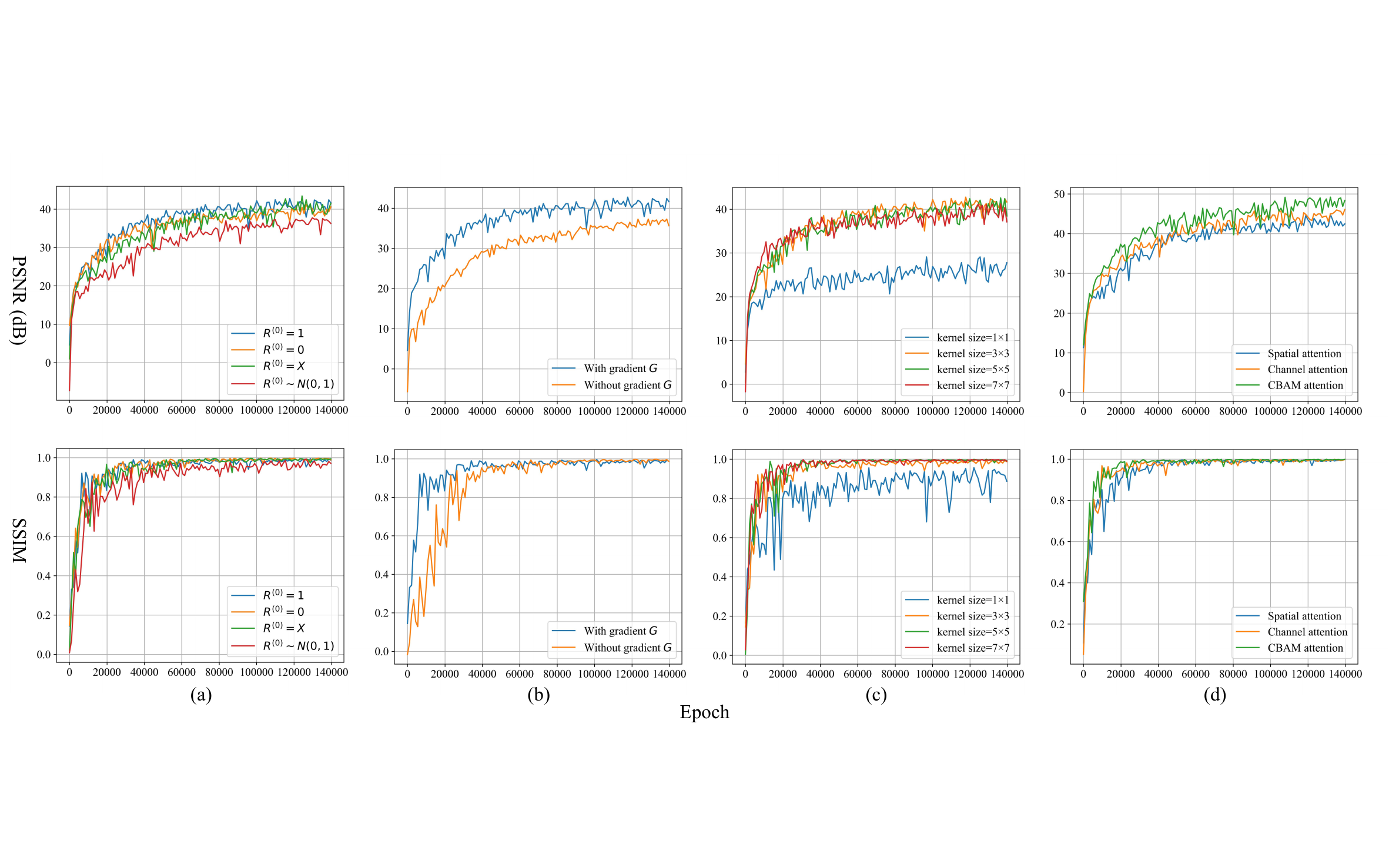}
	\caption{Convergence of PSNR and SSIM corresponding to different ablation experiments.}
	\label{fig4}
\end{figure*}
\subsection{Comparative Experiment}
To highlight the superiority of the proposed method, we selected state-of-the-art and representative deep learning watermarking methods for comparison, including HiDDeN \cite{Zhu2018HiDDeNHD}, StegaStamp \cite{tancik2020stegastamp}, MBRS \cite{jia2021mbrs}, CIN \cite{ma2022towards}, ARWGAN \cite{huang2023arwgan}, Document \cite{ge2023robust}, De-END \cite{9956019} and ST-DCN \cite{ma2025geometric}. We utilized the publicly available pre-trained models provided by these methods and express our gratitude for their open-source contributions.

Table \ref{tab:booktabs3} presents the visual quality and robustness against noise attacks for watermarked images generated by different methods. In the table, entries marked with an asterisk indicate results derived from the original papers. As shown, our method achieves superior robustness when the embedding strength is set to $\alpha=1.0$, with an overall average accuracy exceeding 99.90\%, outperforming existing methods. In terms of visual quality, our method achieves higher PSNR and SSIM values compared to HiDDeN, StegaStamp, and ARWGAN, while being slightly lower than MBRS.

These experimental results demonstrate the exceptional overall performance of our proposed method, primarily attributed to the design of the encoder structure. By employing a parallel architecture, the robustness against each type of noise attack can be independently optimized. Additionally, the adaptive embedding strategy further ensures the visual quality of the generated watermarked images. In summary, our method achieves simultaneous optimization of both visual quality and robustness.
%The experimental results demonstrate the exceptional performance of our proposed method, which can be attributed to the design of the encoder structure. By employing a parallel framework, the robustness against various types of noise attacks can be independently optimized. Additionally, the use of an adaptive embedding strategy further ensures the visual quality.

\subsection{Ablation Experiment}
This section primarily conducts ablation experiments on the proposed method, focusing on the initial iteration state, adaptive embedding strength, receptive field, and attention mechanism. Throughout these experiments, all network parameters and configurations remain consistent and unchanged, except for the specific parameters or local blocks being ablated.
\subsubsection{Initial Iteration State}
Figure \ref{fig4} (a) illustrates the convergence of visual quality when four different sampling spaces are used as the initial iterative optimization targets, while Table \ref{tab:booktabs4} presents the corresponding robustness of the networks. Experimental results demonstrate that our network can optimize from any initial state, achieving both high visual quality and robustness upon convergence. As theoretically analyzed in Section \ref{sec_RIO}, the adopted iterative residual optimization enables the network to progressively approach the neighborhood of optimal residuals from arbitrary initial states through loss minimization, thereby satisfying all constraints including visual quality and robustness requirements.

\subsubsection{Image Gradient}
Figure \ref{fig4} (b) illustrates the convergence of visual quality with and without using image gradients as the adaptive embedding strength for watermarks, while Table \ref{tab:booktabs4} presents the corresponding robustness of the networks. The results demonstrate that leveraging image gradients as adaptive embedding strength significantly enhances both visual quality and robustness. The underlying rationale is straightforward: without adaptive embedding, the amount of watermark is uniformly distributed across all spatial locations of an image, making it difficult to avoid embedding in smooth regions. In contrast, adaptive embedding effectively addresses this issue by concentrating watermark insertion in perceptually complex regions.

\subsubsection{Receptive Field}
In addition, we evaluated the impact of the receptive field size of the network parameters. Figure \ref{fig4} (c) and Table \ref{tab:booktabs4} demonstrate the visual quality convergence and robustness, respectively. The results indicate that a receptive field size of $3\times3$ achieves the optimal performance. A smaller receptive field fails to adequately optimize both visual quality and robustness. In contrast, a larger receptive field increases the number of trainable parameters, making the network more prone to overfitting.

\subsubsection{Attention Mechanism}
Finally, we evaluated the performance of different attention mechanisms, including channel attention (CA), spatial attention (SA), and convolutional block attention module (CBAM) \cite{woo2018cbam}. Given that the encoder primarily generates a robust residual $R$, the attention mechanisms were applied specifically to $R$. The results are presented in Figure \ref{fig4} (d) and Table \ref{tab:booktabs4}. As observed, incorporating attention mechanisms can improve the visual quality of the generated watermarked images. However, robustness against certain types of attacks may be compromised. Therefore, in our proposed method, excellent performance can still be achieved without the use of attention mechanisms.

\section{Conclusion}
To address the challenges of visual quality, robustness, and generalization in deep learning watermarking methods, we propose an adaptive robust iterative watermarking framework. Specifically, we develop a robust iterative watermarking scheme and design an encoder structure to generate watermarked images with strong robustness. Additionally, we leverage image gradients to determine the embedding strength at each pixel, further enhancing the visual quality of the watermarked images. Extensive experiments demonstrate the robustness of our method against various noise attacks and its generalization capability across datasets, while maintaining high imperceptibility in the generated watermarked images. Furthermore, ablation studies validate the effectiveness of our network design. In summary, our watermarking framework significantly improves visual quality and robustness, charting a promising path for future advancements.

%\section*{Acknowledgments}
%This work is supported by the National Natural Science Foundation
%of China (No. 62261160653, No. 62441237, No. 62172435 and No. U23A20305), the %Guangdong Provincial Key Laboratory of Information Security Technology (No. %2023B1212060026).

\bibliographystyle{ACM-Reference-Format}
\bibliography{sample-base}

\appendix

\end{document}